\documentclass[twoside,11pt]{article}

\usepackage{jmlr2e}

\usepackage{lipsum}
\usepackage{amsfonts}
\usepackage{amsmath}
\usepackage{epstopdf}
\usepackage{algorithmic}
\usepackage{algorithm}
\usepackage{caption}
\usepackage{subcaption}
\usepackage{graphicx}

  \newcommand{\bv}[1]{{\bf #1}}

\ShortHeadings{A Locally Adapting Technique for Boundary Detection using Image Segmentation}{Howard, Hock, Meehan, and Dresselhaus-Cooper}
\firstpageno{1}

\begin{document}

\title{A Locally Adapting Technique for Boundary Detection using Image Segmentation}

\author{\name Marylesa Howard \email howardmm@nv.doe.gov \\
       \addr Signal Processing and Applied Mathematics\\
       Nevada National Security Site\\
       P.O.~Box~98521, M/S NLV078, Las Vegas, NV 89193-8521, USA
       \AND
       \name Margaret C. Hock \email hockmc@nv.doe.gov \\
       \addr Signal Processing and Applied Mathematics\\
       Nevada National Security Site\\
       P.O.~Box~98521, M/S NLV078, Las Vegas, NV 89193-8521, USA\vspace{5pt}\\
       Department of Mathematical Sciences\\
       University of Alabama in Huntsville\\
       301~Sparkman~Drive, 258A, Huntsville, AL 35899, USA
       \AND
       \name B. T. Meehan \email meehanbt@nv.doe.gov\\
       \addr Signal Processing and Applied Mathematics\\
       Nevada National Security Site\\
       P.O.~Box~98521, M/S NLV078, Las Vegas, NV 89193-8521, USA
       \AND
       \name Leora Dresselhaus-Cooper \email lacooper@mit.edu\\
       \addr Department of Physical Chemistry, Institute for Soldier Nanotechnology\\
       Massachusetts Institute of Technology\\
       77~Massachusetts~Avenue, NE47-593, Cambridge, MA 02139, USA}

\editor{To Be Determined}

\maketitle

\begin{abstract}
Rapid growth in the field of quantitative digital image analysis is paving the way for researchers to make precise measurements about objects in an image. To compute quantities from an image such as the density of compressed materials or the velocity of a shockwave, object boundaries must first be determined, where images containing regions that each have a spatial trend in intensity are of particular interest here. We present a supervised, statistical image segmentation method that incorporates spatial information to locate boundaries between regions with overlapping intensity histograms. The segmentation of a pixel is determined by comparing its intensity to distributions from nearby pixel intensities. Because of the statistical nature of the algorithm, we use maximum likelihood estimation to quantify uncertainty about each boundary. We demonstrate the success of this algorithm at locating boundaries and providing uncertainty bands on a radiograph of a multicomponent cylinder and on an optical image of a laser-induced shockwave.
\end{abstract}


\section{Introduction} 
Quantitative digital image analysis goes beyond qualitatively discriminating between materials or phenomena in a scene and instead allows for precise measurements to be made about objects. Boundary detection is a vital part of extracting information encoded in images, allowing for the computation of quantities of interest including density, velocity, pressure, etc. For example, \citet{Moriwaki} calculates the volume of fruit by identifying fruit boundaries from magnetic resonance images and \citet{Byng} monitors changes in breast tissue location and density to infer breast cancer risk. 

We present a supervised image segmentation method that is designed to identify boundaries for images with spatially varying intensity, including those with high noise and low contrast. As the purpose of this method is boundary identification, we assume that the user will provide training data for a majority of the image. To assign each pixel, our algorithm considers only locally occurring classes to build a statistical segmentation model. The user defines \textit{local} with two parameters that reduce the amount of training data considered for the identification of the pixel's class. Our algorithm determines the most probable class for each pixel by comparing its intensity with intensity distributions for each class present in the local training data. 

To enhance its utility for applications, we combine our method for boundary identification with a statistical analysis to determine uncertainty bands about identified boundaries to provide additional quantitative information describing the objects in the image. In this work, we measure the uncertainty in the segmented boundary using two different spatially resolved metrics. Because of the statistical nature of our algorithm, we employ maximum likelihood estimation to quantify uncertainty about each boundary and analysis of variance to identify the separability of classes at the boundaries. Both analyses provide pixel-by-pixel tests that produce uncertainty (p-value) maps for the segmentation. The boundary uncertainties determined in this way can then be propagated through calculations, giving error bars on quantities. We provide a novel method to objectively obtain statistically justified error bars on image quantities that are measured by segmentation boundary identification.

\subsection{Background}
Computational boundary detection methods typically fall into two main categories: edge detection and segmentation. Classical edge detection methods include the Sobel edge detector \citep{Kittler} and gradient methods \citep{Haralick} and are often most successful when distinguishing between regions that have high contrast \citep{Mansoor}. Image segmentation determines boundaries between regions by partitioning an image into separate classes or materials \citep{Hastie,Pal}. 
Advancements in computing technology have enabled the development of many sophisticated segmentation and edge detection methods, ranging in computational and mathematical complexity, including statistical discriminant analysis models \citep{Hastie}, support vector machines that downselect training data \citep{Bishop}, fuzzy logic edge detection \citep{Melin}, gradient-based image segmentation \citep{Hell}, and deep convolutional neural networks \citep{Xie}. 

Many advanced methods are developed to solve specific application problems, yet the value of each method lies within the particular data sets used and the experience of the analyst to extract the best performance from each method \cite{Hand}. In fact, to compute quantities from images, communities within  medicine and remote sensing manually determine object boundaries due to image qualities such as low contrast, heteroskedasticity, and objects whose intensities vary spatially \citep{Bazille,Plaza,Zha}. Manual analysis is time consuming, and makes it difficult to determine mathematically justified errors and uncertainties on manual segmentations.

In addition to identifying class and boundary locations, applications benefit from understanding associated uncertainties, with typical measures of error including k-fold crossvalidation, confusion matrices, and Kappa statistics. These statistical measures provide an overall assessment of the analysis, but recent literature highlights the need for methods that spatially identify uncertainty in image segmentation \citep{Woodcock,Zhao}. For neural networks, \citet{Carpenter} produces a confidence map that indicates the number of voting networks that agree on each pixel's predicted label and \citet{McIver} presents a method for estimating a pixel-scale confidence map when using boosting. Work by \citet{Khatami} takes an alternative approach to developing uncertainty maps by using the spectral domain rather than the spatial domain. \citet{Bogaert} presents an information-based criterion for computing a thematic uncertainty measure that describes the overall spatial variation of the segmentation accuracy.\\

The paper is organized as follows. Our locally-adaptive image segmentation algorithm is detailed in Section \ref{sec:lada}, with corresponding uncertainty quantification for boundary identification in Section \ref{sec:uq}. Numerical results for two real world images are presented in Section \ref{sec:experiments}, and the conclusions follow in Section \ref{sec:conclusions}.

\section{Locally adaptive discriminant analysis}
\label{sec:lada}

\begin{figure}
(a)\includegraphics[width = 0.55\textwidth]{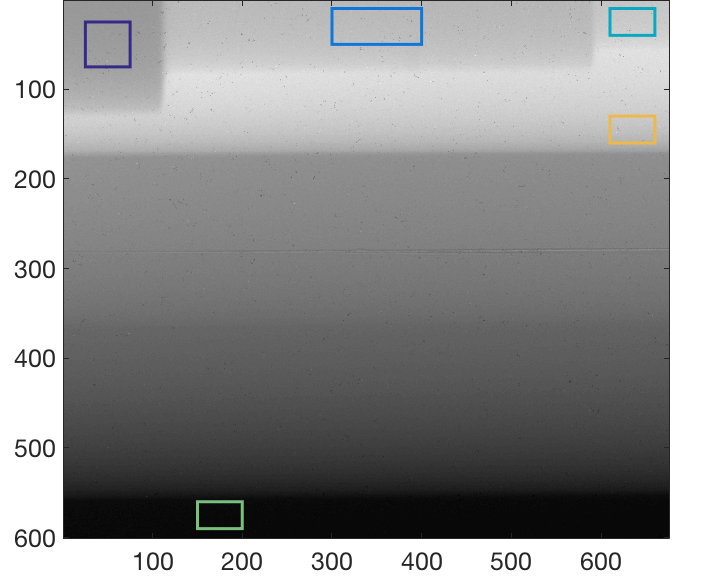}
(b)\raisebox{0.25\height}{\includegraphics[width=.35\textwidth]{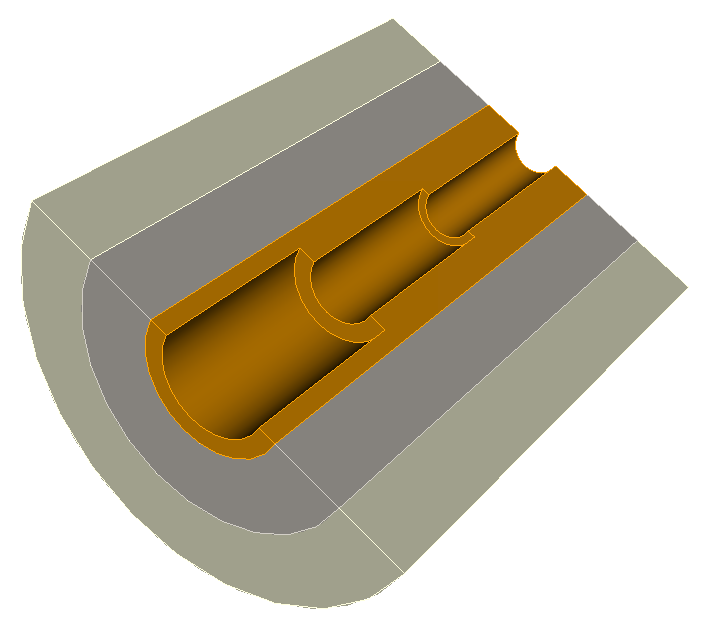}}\\
\begin{center}(c)\includegraphics[width=0.75\textwidth]{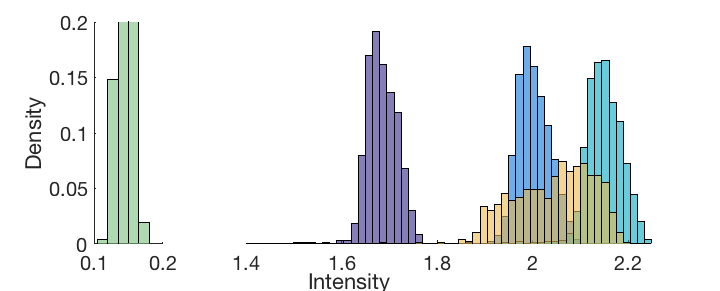}\end{center}
\caption{(a) A radiograph of half of a radially symmetric cylinder calibration object, with the center of the cylinder at the top of the image. The cylinder is comprised of concentric materials: copper, aluminum, and teflon, with varying bore widths in its center (air) that are shown at the top of the image, as well as an area beyond the cylinder (air) at the bottom of the image. The purple, blue and green rectangles indicate areas that all belong to the same class: air, and the yellow rectangle indicates an area belonging to the copper class. (b) A graphic representation of the cylinder. (c) Histograms of the regions outlined in (a); note there is a break in the $x$-axis.}
\label{fig:abel_globalVsLocal}
\end{figure}

Global image segmentation methods do not account for spatial variation in intensity within a class. The mean and variance of a class may vary spatially within an image. By restricting our knowledge to local information about a pixel, we will have a better sense of the pixel's true class, without being misled by information on the other side of the image. For example, Figure \ref{fig:abel_globalVsLocal}(a) is a radiograph from the Nevada National Security Site of a radially symmetric cylinder turned on its side, with a graphic representation given in Figure \ref{fig:abel_globalVsLocal}(b). The cylinder is hollow in its center (top of image (a)) with varying bore widths, and the adjacent concentric layers are copper, aluminum, and teflon, followed by air outside the cylinder (bottommost region in image (a)). The varying thicknesses of the materials affect the intensity on the radiograph as it is a function of areal density, and an objective in the quantitative analysis of this image is to identify boundaries between materials in order to compute material density \citep{Howard16}. The purple, blue, teal, and green rectangles in image (a) all represent the air class, and the intensity of the pixels contained in those colored rectangles are shown in the respectively colored histograms given in Figure \ref{fig:abel_globalVsLocal}(c). While each of those regions of pixels come from the same class, they are completely separated in feature space, indicating a spatial relationship between intensity and class. Furthermore, the yellow rectangle in image (a) corresponds to a region of copper, and its corresponding intensity values are plotted in image (c). The copper intensities completely overlap with two of the air intensity histograms, suggesting that a global approach to image segmentation may lead to issues with non-separable data. 

We present our method, locally adaptive discriminant analysis (LADA), which restricts the training data for each pixel of interest to the locally-occurring classes based on two user-selected parameters and builds a trainer based on the Gaussian assumption of discriminant analysis. 
LADA is, therefore, a local implementation of discriminant analysis to adequately separate classes with spatially-varying intensities and is appropriate for images with shadows, with heterogeneous illumination, or of areal density, including those with ow contrast and/or high noise. In addition to the user-defined classes, LADA allows for a \textit{bonus class} to be selected for a pixel when there are not enough, or no, local training data to represent a class. The algorithm is described here, with pseudocode in Algorithm~\ref{alg:LADA}.

Given an image $X$, a pixel $x_{ij}$, and the set of training data $T\subset X$, we are interested in determining the most likely class to which $x_{ij}$ belongs, for each $x_{ij}\in X$. The classes, $\omega_c$, are defined by the training data, with $c=1,\ldots,C$, for a total of $C$ known classes occurring in $X$. Let $T_{\omega_c} \subset T$ be the set of all training pixels for class $\omega_c$.

Rather than considering the entire set $T$ to build a trainer for $x_{ij}$, we reduce the training data via two user-selected parameters: $d$ and $n$. Given the distance parameter $d$, we define the subimage $S_{ij} \subset X$ about pixel $x_{ij}$ to be 
\begin{equation*}
 S_{ij} =  \left\{  x_{kl} \; \Big{|} \; \sqrt{(i-k)^2+(j-l)^2} \leq d \right\},
\end{equation*}
such that $S_{ij}$ is the set of all pixels within radius $d$ to pixel $x_{ij}$. All known regions (training data) outside $S_{ij}$ are temporarily ignored and will have no effect on the segmentation of $x_{ij}$. Further reducing the trainer's view of \textit{local}, for each class $\omega_c$, we define the local training data, $T_{\omega_{c}{ij}}\subset S_{ij}$, to be the set of, at most, $n$ nearest (Euclidean distance) training pixels to $x_{ij}$, where ties are broken via lexicographical ordering. Together, $d$ and $n$ serve as our definition of \textit{local} for LADA.

\begin{figure}\begin{center}
\includegraphics[width=5.5in]{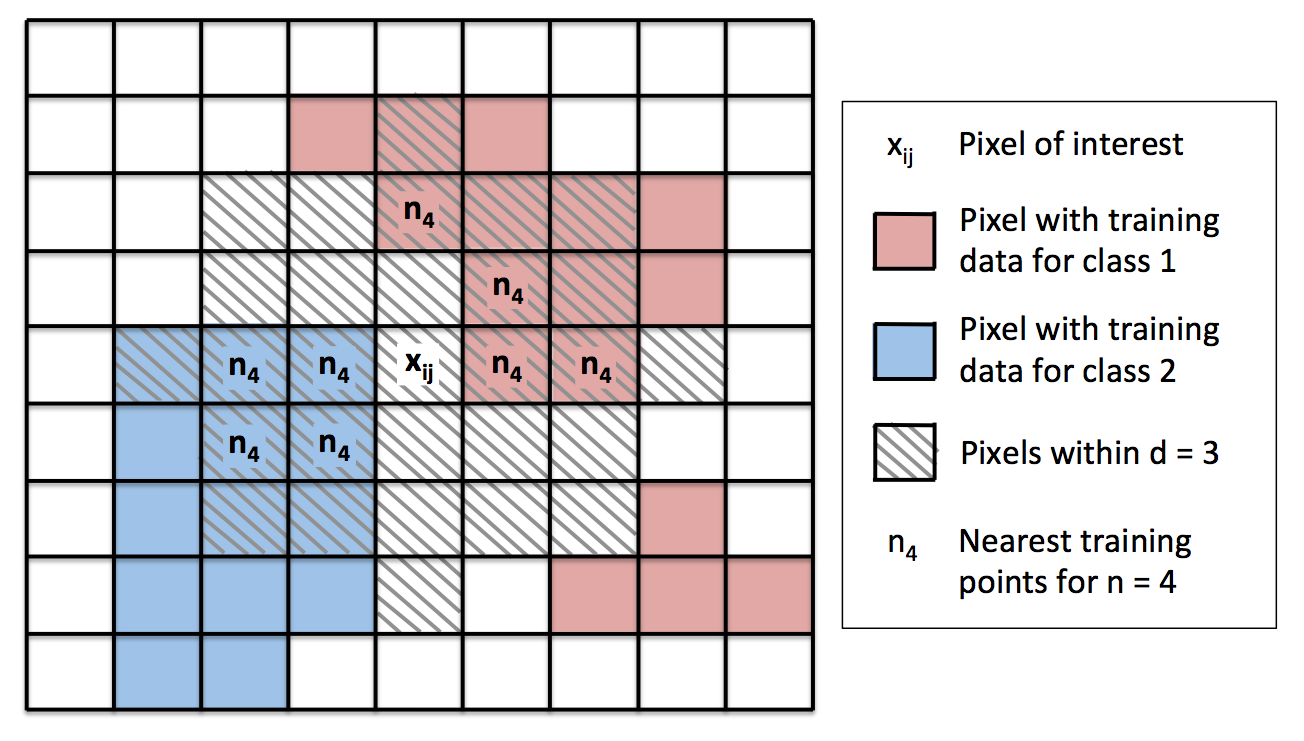}
\caption{An example of defining \textit{local} training data with $d=3$ and $n=4$. There are two classes with training data represented by pink and blue pixels. The pixels within $S_{ij}$ are indicated with diagonal gray lines, and the nearest four training points for each class are marked with $n_4$.}
\label{fig:LADAparameters}\end{center}
\end{figure}

A simplified visual example of the restriction of the local training data for parameters $d=3$ and $n=4$ is given in Figure \ref{fig:LADAparameters}. The two training data classes that make up $T$ are shown in pink (class 1) and blue (class 2) colored pixels, with the center pixel identified as $x_{ij}$. All pixels in subimage $S_{ij}$ (within radius $d=3$) are shaded, with elements of $T_{\omega_1{ij}}$ being the pink, shaded pixels with demarcation of $n_4$ and elements of $T_{\omega_2{ij}}$ being the shaded, blue pixels with demarcation of $n_4$. Notice this selection of four nearest training points for the blue class is not unique but obeys the lexicographical ordering of left to right, top to bottom.

Each class $\omega_c$ with $|T_{\omega_c{ij}}|\geq3$ is considered a potential class for $x_{ij}$ such that we can compute statistics on the locally occurring training data. A set with $|T_{\omega_c{ij}}|<3$ is possible even with $n\geq 3$ if not enough training pixels of class $c$ occur within $S_{ij}$, and, in such a case, that class is not considered within the computations for \textit{local}. We assume the local training data for class $\omega_c$ are drawn from a Gaussian distribution, with mean $\mu_{\omega_c{ij}}$ and standard deviation $\sigma_{\omega_c{ij}}$. By restricting to local training data, a Gaussian distribution assumption is often more reasonable than on a global scale where the full training data set often violates such an assumption in images with spatially-varying intensities. This will be demonstrated in Section \ref{sec:experiments}.

The class $\omega_c$ to which $x_{ij}$ most likely belongs, given these distributional assumptions, is defined to be
\begin{equation*}
G(x_{ij}) =  \underset{\omega_c}{\arg} \left\{p(x_{ij}|\mu_{\omega_c{ij}},\sigma_{\omega_c{ij}}) > p(x_{ij}|\mu_{\omega_b{ij}},\sigma_{\omega_b{ij}}) \; \forall \; \omega_b\neq \omega_c\right\},
\end{equation*}
where, without loss of generality,
\begin{equation*}
x_{ij}\sim \mathcal{N}(\mu_{\omega_c{ij}},\sigma_{\omega_c{ij}}),
\end{equation*}
and $G(\cdot)$ is the function mapping a pixel into its segmented class. 

It is possible that, via choice of $d$, there are too few training points within subimage $S_{ij}$ to reliably compute a standard deviation (i.e., $|T_{\omega_c{ij}}|\leq2, \; \forall \; c$). In such a case, we place $x_{ij}$ into the \textit{bonus class}, 
\begin{equation*}
G(x_{ij}) = C+1,
\end{equation*}
indicating there was not enough local information to identify to which of the $C$ classes it belongs. In general, if a significant portion of the image is being placed into the bonus class, the analyst might consider choosing more training data if more are known or choosing a larger distance parameter $d$.

Furthermore, it should be noted that as $d$ approaches the bounds of the diagonal distance of the image and $n$ is increased to the magnitude of the training data sets for each class, the effect of looking at local training data diminishes and the focus becomes global. In such a case, this algorithm simply becomes quadratic discriminant analysis \citep{Hastie}. The algorithm may be restricted further to perform similarly to linear discriminant analysis if $\sigma_{\omega_c{ij}} = \sigma_{{ij}}$ for all considered classes $\omega_c$.

\begin{algorithm}
\caption{Locally adaptive discriminant analysis}
\label{alg:LADA}
\vspace{.05in}

Given image $X$, define training data $T$ with $C$ classes, and \textit{local} parameters $d$ and $n$.
\vspace{.05in}

For each pixel $x_{ij}\in X$:
\begin{enumerate}
\item {Define $S_{ij}$ to be the subimage of $X$ centered at $x_{ij}$ with radius $d$ such that $$S_{ij} =  \left\{  x_{kl} \; \Big{|} \; \sqrt{(i-k)^2+(j-l)^2} \leq d \right\}.$$}
\item {For $c=1$ to $C$:}
\begin{itemize}
\item[a.] {Define the local training data $T_{\omega_c{ij}}$ to be the set of the $n$ nearest training points to $x_{ij}$ within $S_{ij}$ that belong to class $\omega_c$, with ties broken via lexicographical order.}
\item[b.] {Compute the mean $\mu_{\omega_c{ij}}$ and standard deviation $\sigma_{\omega_c{ij}}$ of $\{T_{\omega_c{ij}}\}$.}
\end{itemize}
\item {Place $x_{ij}$ into class $\omega_c$ for which
$$G(x_{ij}) = \left\{ \begin{array}{ll} \arg\max\limits_c p(x_{ij}|\mu_{\omega_c{ij}},\sigma_{\omega_c{ij}}) & \text{if } |T_{\omega_c{ij}}| \geq 3\\ C+1 & \text{otherwise}\end{array}\right\},\quad \text{where }x_{ij}\sim \mathcal{N}(\mu_{\omega_c{ij}},\sigma_{\omega_c{ij}}).$$}
\end{enumerate}
\end{algorithm}

For images with strong spatial variation even in local regions, the values of $d$ and $n$ should be chosen with smaller magnitudes if significant training data can be provided to avoid abundant bonus class assignment. There can be a tradeoff between a priori knowledge and the selection of parameters $d$ and $n$. Since LADA is an edge detection method, we assume the majority of the image is known, sans the boundaries.

\section{Uncertainty on boundary detection}
\label{sec:uq}

We build two uncertainty maps corresponding to the segmentation that describe, first, our confidence in the selected segmentation based on local information, and second, our ability to discriminate between classes at boundaries. For the former, we apply maximum likelihood estimation, and, for the latter, we apply a standard analysis of variance (ANOVA) test.

\subsection{Maximum likelihood estimation p-value}
\begin{figure}\begin{center}
(a)\includegraphics[width=0.45\textwidth]{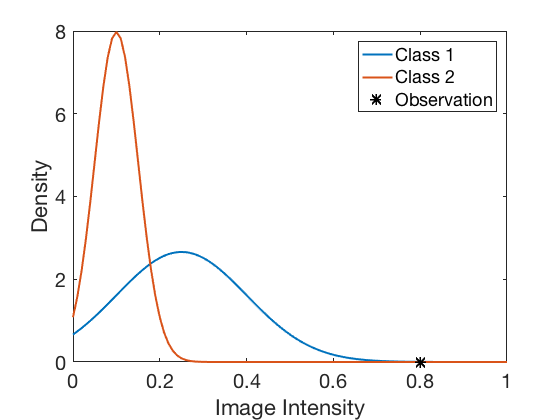}
(b)\includegraphics[width=0.45\textwidth]{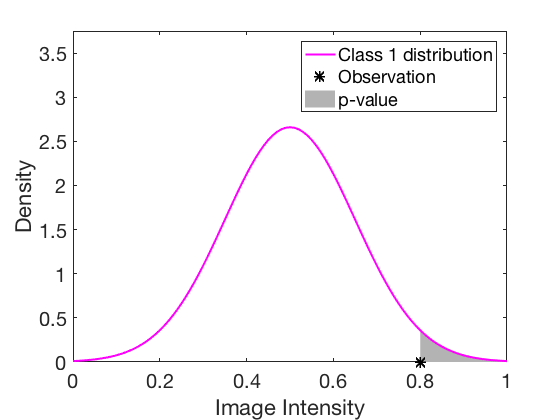}
\caption{(a) While every pixel is segmented as belonging to a class, the assigned class is not always particularly likely. For the observation (black star), neither class 1 nor class 2 is very probable, though it is more likely to belong to class 1 (blue line) based on probability density. (b) For a given class distribution (magenta line) and observation (black star), the p-value is computed as the probability of observing that value or something more extreme.}
\label{fig:dists}\end{center}
\end{figure}

Given a LADA segmentation for a pixel, we wish to quantify the probability that the pixel belongs to that class. Consider the example in Figure \ref{fig:dists}(a): given the Gaussian distributions for two, hypothetical local classes and the intensity observation marked by the black star, the pixel would be segmented into class 1, given by the blue line, since it has greater probability density at that observation. However, because the observation is on the tail of the class 1 distribution, it is not well represented by that class either. 

The estimation method by which LADA determines the segmented class, $c$, is considered a maximum likelihood estimator (MLE). In terms of segmentation, MLE is a technique for determining the class that maximizes the probability distribution for an observed pixel intensity \citep{DeGroot}. For a grayscale image, we assume the local population mean $\mu_{\omega_c{ij}}$ and variance $\sigma_{\omega_c{ij}}$ are known for the class $\omega_c$ to which pixel $x_{ij}$ is segmented, obtained via the local training data. The class was selected using the optimization
\begin{equation*}
\widehat{\omega_c} = \underset{\omega_c}{\arg}  \max\limits_{(\mu_{\omega_c{ij}},\sigma_{\omega_c{ij}})} f(x|\mu_{\omega_c{ij}},\sigma_{\omega_c{ij}}),
\end{equation*}
where 
\begin{equation*}
x \sim \mathcal{N}(\mu_{\omega_c{ij}},\sigma_{\omega_c{ij}}).
\end{equation*}
From this, we compute the p-value, the probability of observing the pixel $x_{ij}$ or something more extreme, given the segmented class $\omega_c$ and its associated parameters:
\begin{equation*}
p = \left\{\begin{array}{ll}
\text{P}\left(X \geq x_{ij}|\mu_{\omega_c{ij}},\sigma_{\omega_c{ij}}\right) & \text{if } x_{ij}>\mu_{\omega_c{ij}}\\
\text{P}\left(X \leq x_{ij}|\mu_{\omega_c{ij}},\sigma_{\omega_c{ij}}\right) & \text{otherwise}
\end{array}\right\}.
\end{equation*}

Assuming pixel $x_{ij}$ had equal probability of having observed either positive or negative noise, we multiply the p-value by two, for a two-sided p-value, which is demonstrated visually in Figure \ref{fig:dists}(b) for a distribution with mean $\mu_{\omega_c{ij}} = 0.5$ and $\sigma_{\omega_c{ij}} = 0.38$, and observation $x_{ij} = 0.8$. For an image segmented by LADA, we can produce a corresponding image of p-values from the statistical inference given here. This concept can be extended to multibanded images (non-grayscale) using a multivariate normal distribution on the colored pixel $\bv{x}_{ij}$.

\subsection{Analysis of variance p-value}
Analysis of variance (ANOVA) provides a hypothesis test as to whether local classes are equal in mean intensity or not by analyzing the variance of the local classes \citep{Casella}. We model the random variable pixel intensity as
\begin{equation*}
X_{ij} = \bv{\mu}_{\omega_c{ij}} + \bv{\epsilon}_{\omega_c{ij}},
\end{equation*}
where $X_{ij}$ may be of any length. Thus each pixel from a given class has a known mean, $\bv{\mu}_{\omega_c{ij}}$, and some noise, $\bv{\epsilon}_{\omega_c{ij}}$, which makes up the observed intensity. Assuming the $\bv{\epsilon}_{\omega_c{ij}}$ are iid Gaussian with zero mean, equal variances, and zero covariances, the null hypothesis states that all the class means are equal and the alternative states that at least one mean is different. Rather, we are interested in knowing if any two means of local classes are evidenced to be equal, so we perform multiple ANOVA's to compare only two classes at a time. In our case, the hypotheses of interest are 
\begin{align*}
H_0: & \;\; \bv{\mu}_{\omega_c{ij}} = \bv{\mu}_{\omega_b{ij}},\\
H_a: & \;\; \bv{\mu}_{\omega_c{ij}} \neq \bv{\mu}_{\omega_b{ij}},
\end{align*}
for all classes $\omega_c$ and $\omega_b$ local to pixel $\bv{x}_{ij}$.

The resulting p-value of the ANOVA test quantifies evidence for or against the null hypothesis. A large ANOVA p-value indicates that we have evidence in favor of the null hypothesis which states that the two classes have the same mean and are undifferentiable with such a measure. We compute all ANOVA paired p-values and consider the largest value, which provides the most evidence that there are at least two classes with equal means, which can be visualized with an image corresponding to each pixel of the original image. Thus, the ANOVA image is used to indicate ``problem areas'' within an image that has data which are especially difficult to discriminate between classes, or may be improved upon by editing the training data. In the case of multibanded images, multivariate analysis of variance (MANOVA) is used.

\section{Numerical results}
\label{sec:experiments}
In this section, we include results on grayscale images from two physics applications. In demonstrating the utility of LADA, we highlight issues seen in images typical to different fields of science that pose problems for conventional segmentation methods. The first image is a radiograph of a static, multi-layered cylindrical object. The second image is of a laser-induced converging shockwave, collected using optical imaging \citep{Pezeril}.

\subsection{Static data: multi-material cylinder}

\begin{figure}
\begin{center}
(a)\includegraphics[width = 0.65\textwidth]{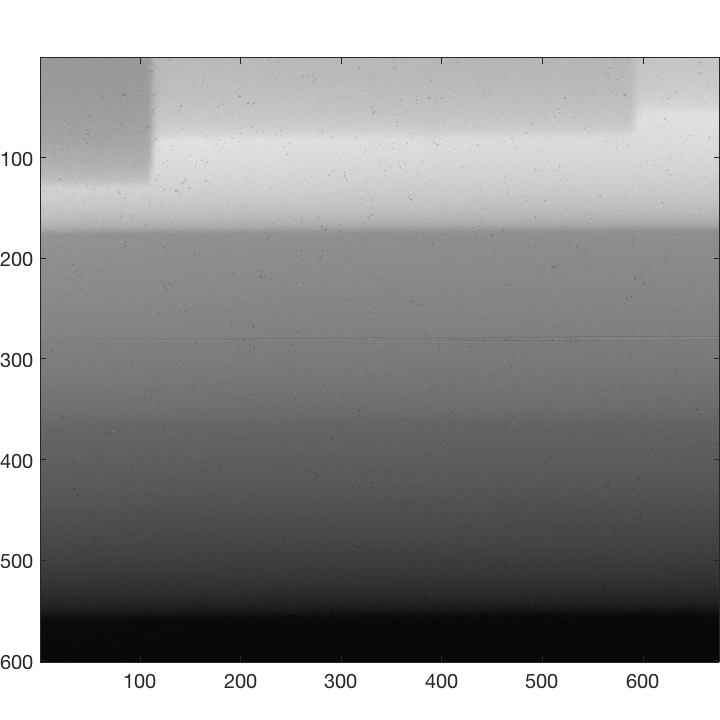}\\
(b)\includegraphics[width = 0.45\textwidth]{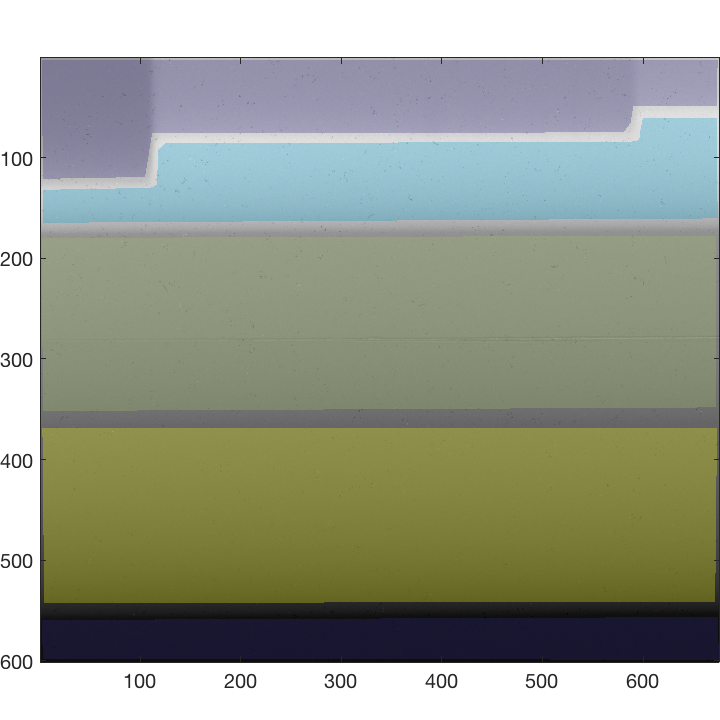}
(c)\includegraphics[width = 0.45\textwidth]{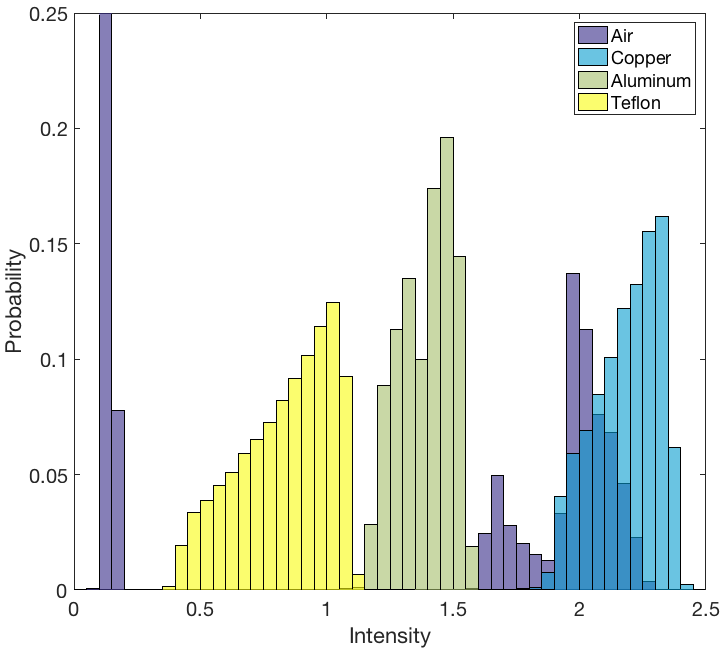}
\caption{A radiograph of a partial cylinder (a). Training data for the cylinder are superimposed on the image in (b). The histogram of global training data (c) demonstrates the non-Gaussian distributions of the classes and the trimodal nature of the air class.}
\label{fig:abel}\end{center}
\end{figure}

A radiograph from the Cygnus X-ray machine at the U1a underground facility at the Nevada National Security Site is shown in Figure \ref{fig:abel}(a), and was presented in Section \ref{sec:lada}. It is half of a radially symmetric cylinder with the axis of rotation at the top of the image. The cylinder is hollow, with varying bore widths, and has concentric cylinders of copper, aluminum, and teflon. The final material (darkest horizontal layer in the image) is outside the cylinder, i.e. air. A prime objective in the quantitative analysis of this image is to identify boundaries between materials for subsequent calculations of material density.

Boundary identification for the concentric cylinders is made difficult by the intensity gradient across each class corresponding to a change in areal density. The training data are provided in Figure \ref{fig:abel}(c), superimposed on the radiograph, and comprise $88\%$ of the image. A large amount of training data is provided to accommodate the locally-adaptive nature of LADA, which requires training data throughout the image. It is simple to provide a large quantity of training data since the only aspect of the image that is unknown is the boundaries. Classical segmentation methods have difficulty correctly segmenting the class representing air, which has a very wide, multimodal distribution as it appears at the top (light to medium gray) and bottom (black) regions of the image, as shown in the training data histograms in Figure \ref{fig:abel}(d). The histograms demonstrate overlap between class probability densities, which is the driving reason why classical threshold methods like quadratic discriminant analysis (QDA) do not work well on data such as these with spatial gradients in classes. In addition, the Gaussian assumption of QDA is not appropriate for the training data histograms.

\begin{figure}\begin{center}
(a)\includegraphics[width = 0.58\textwidth]{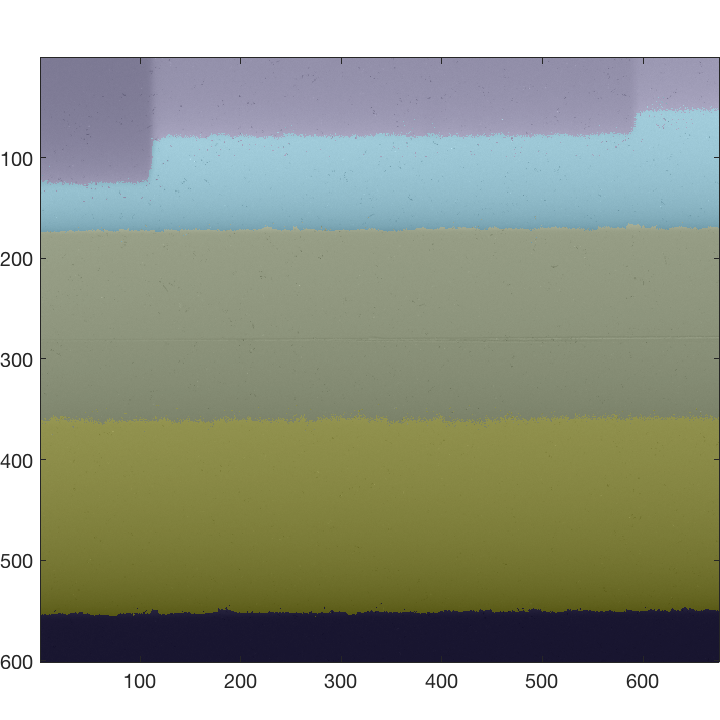}\\
(b)\includegraphics[width = .4\textwidth]{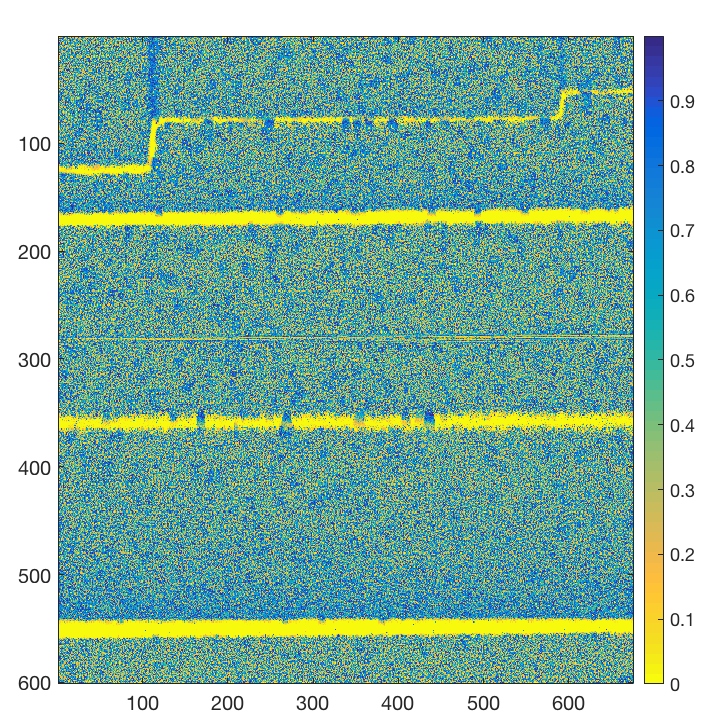}
(c)\includegraphics[width=.4\textwidth]{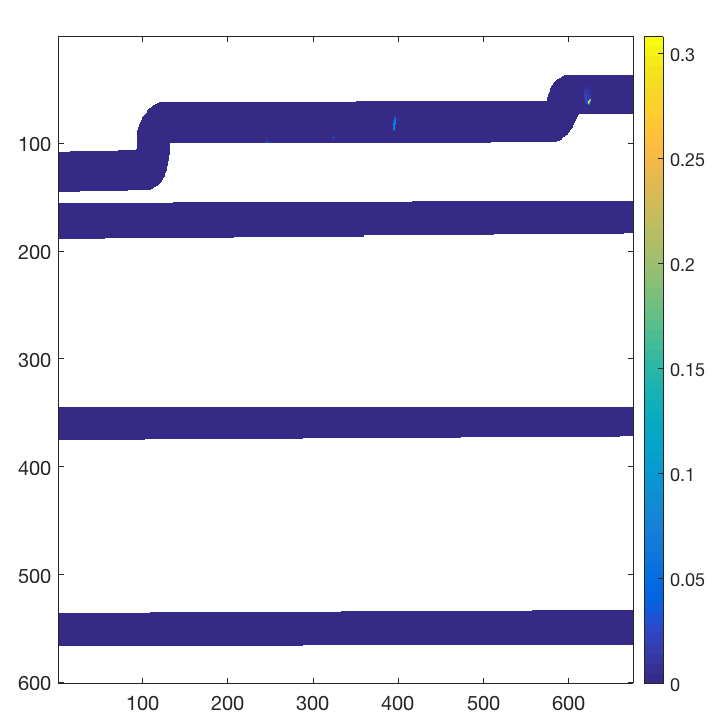}\\
(d)\includegraphics[width = .4\textwidth]{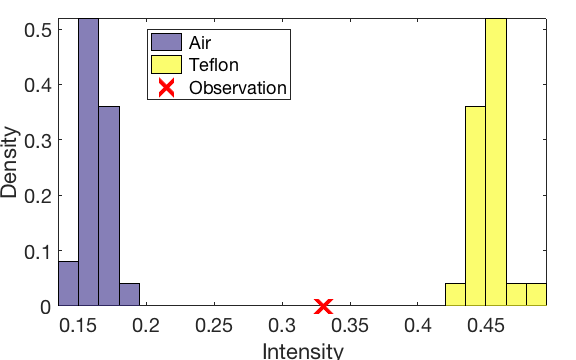}
(e)\includegraphics[width=0.3\textwidth]{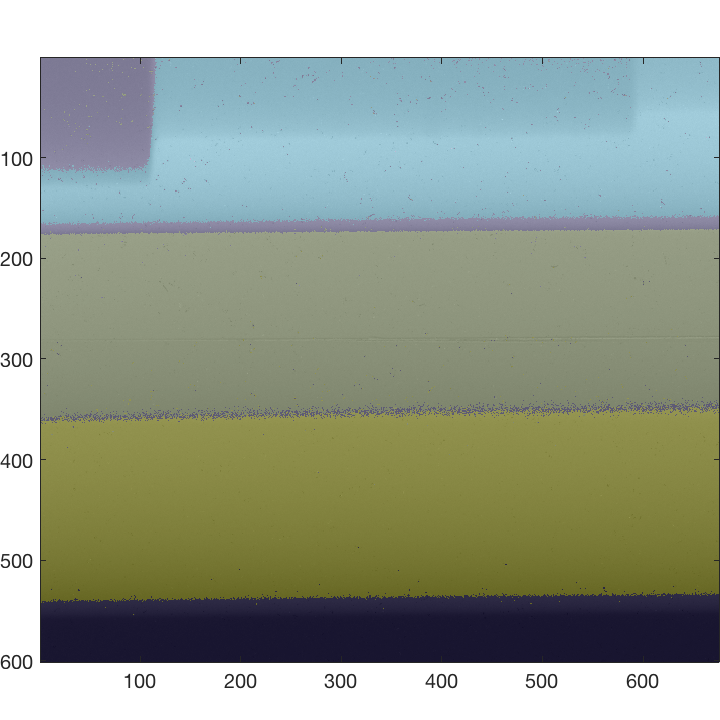}
\caption{The LADA segmentation (a) and associated MLE p-value map (b) and ANOVA p-value map (c) provide insight into boundary location uncertainty and class discrimination, respectively. (d) Histogram of local training data for a boundary pixel. (e) QDA segmentation misrepresenting boundaries.}
\label{fig:abel_seg}
\end{center}
\end{figure}

The LADA segmentation is provided in Figure \ref{fig:abel_seg}(a), based on the training data in Figure \ref{fig:abel}(c), with parameters $d = 25$ and $n=25$. The segmentation captures the large-scale features of three materials, surrounded on either side by air, and has very few ``noisy'' segmented pixels, i.e. a pixel whose class is unlike its immediate neighbors. The implied boundaries between classes are likely smoother in actuality, with the roughness likely due to image quality or parameter selection. Reducing the view of local may help alleviate segmenting errors due to noise.

Figure \ref{fig:abel_seg}(b) visualizes the MLE p-values that describe how probable the chosen class is, given the observed pixel intensity. The bright yellow regions indicate pixels that are not well represented by the local training data in the available classes. The thickness of the yellow lines is influenced by the quality of the image and how close to the boundaries the user provides training data. In this image, having a pixel near a boundary between classes (with the closest training data being many pixels away) can lead to that pixel not looking similar to either training subset. For example, consider a pixel near the boundary of air and teflon at (49,551) in the lower left-hand corner of Figure~\ref{fig:abel_seg}(a) and its corresponding local training data presented in Figure~\ref{fig:abel_seg}(d) [We note the local training data shown here pass the Shapiro-Wilk test for normalcy]. In this case, the closest training data are at least ten pixels away for either class, and the observed pixel is not well represented by either class due to the strong intensity gradient in this region. The yellow regions of the MLE p-value map indicate all of these such circumstances. 

In providing training data, the user implies a level of uncertainty about the boundary location through the width of the void between classes. The regions of uncertainty about the material boundaries provided by LADA (p-value $<$ 0.05 in Figure \ref{fig:abel_seg}(b)) are thinner than the corresponding regions in Figure \ref{fig:abel}(c). For example, the horizontal region of uncertainty centered about the y-axis value of 365 has a width of roughly ten pixels, whereas the corresponding void in training data of Figure \ref{fig:abel}(c) has a width of roughly eighteen pixels. LADA has reduced the a priori user uncertainty by nearly half.

The ANOVA p-value map is given in Figure~\ref{fig:abel_seg}(c). The white regions correspond to areas in which only one local class is present within $S_{ij}$ and are thus far away enough from the boundary such that no other class is considered \textit{local}. The dark blue regions (low p-value) indicate that the local training data of the classes are well separable. Lighter regions, mostly absent in this example, indicate that the local training data for at least two of the classes are not well separable. Thus, our training data demonstrate that the classes at the boundaries are well separable with our definition of local parameters. 

For comparison, the classical segmentation via QDA is provided in Figure \ref{fig:abel_seg}(e). It is visibly clear that the global assumptions of QDA fail to provide physically correct boundaries between classes, and the method misidentifies large regions of the image, completely missing two of the hollow bores at the top center and right regions in the image. In addition, the QDA segmentation incorrectly suggests a higher pitched angle of the cylinders from horizontal than LADA, as if the image has been rotated slightly, and an inclusion of air (purple) between each of the layers.

\begin{figure}
\begin{center}
\includegraphics[width=.9\textwidth]{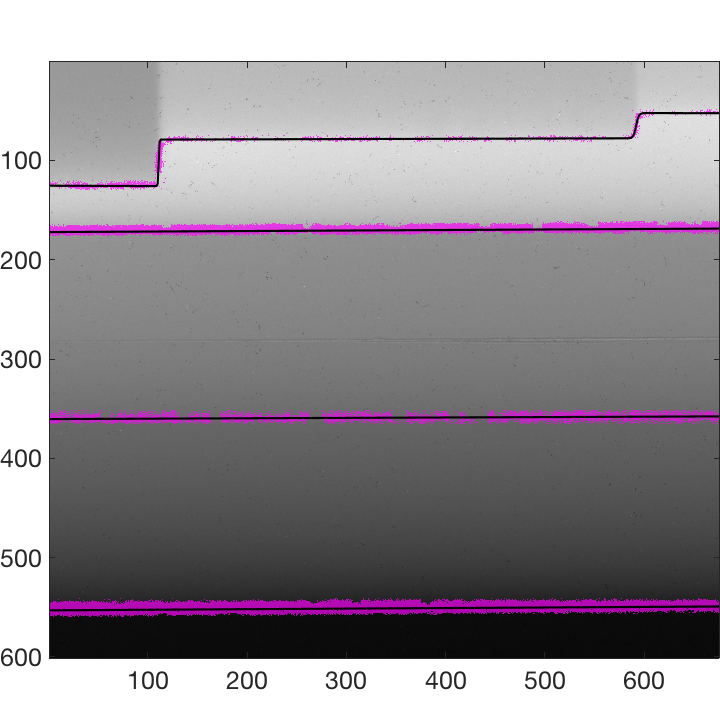}
\end{center}
\caption{The final boundaries (black) with uncertainty bands (purple shade) for the calibration object.}
\label{fig:Abel_UncertWBounds}
\end{figure}

From the segmentation given in Figure \ref{fig:abel_seg}(a), the boundaries between classes are obtained by taking the gradient of the segmentation and fitting lines to the horizontal boundaries and logistic curves to the vertical boundaries, which are reasonable assumptions as we know the geometry of the calibration object. The final, fitted boundaries are given by the black lines in Figure \ref{fig:Abel_UncertWBounds}. To compute regions of uncertainty about the boundaries, we choose a significance level of $\alpha=0.05$ on the p-value map from Figure \ref{fig:abel_seg}(b). All pixels with a p-value $<\alpha$ that are nearby a detected boundary are highlighted in purple in Figure \ref{fig:Abel_UncertWBounds}, indicating a region of uncertainty. Note that the regions of uncertainty are not required to be centered about each boundary. 

\subsection{Dynamic data: cylindrically converging shock waves}

\begin{figure}
\begin{center}
(a)\includegraphics[width=0.45\textwidth]{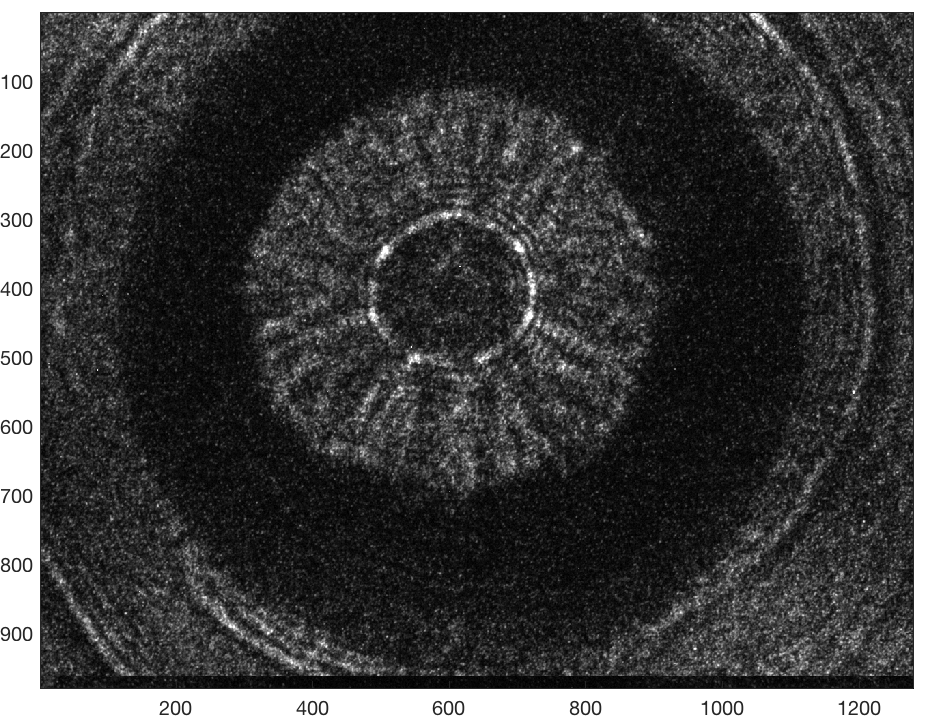} 
(b)\includegraphics[width=0.45\textwidth]{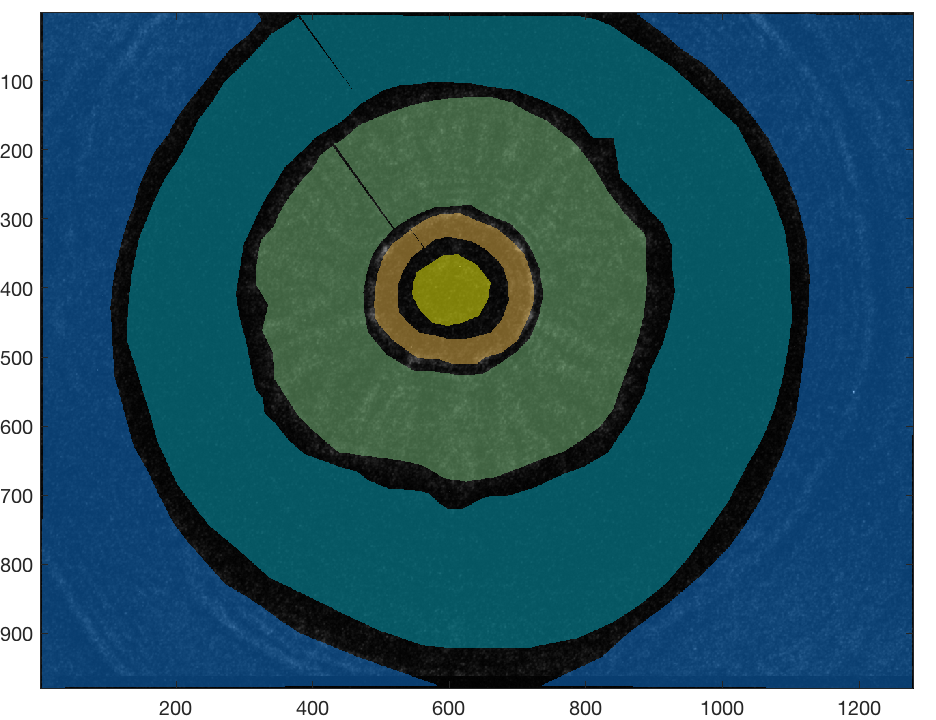} \\
(c)\includegraphics[width = 0.65\textwidth]{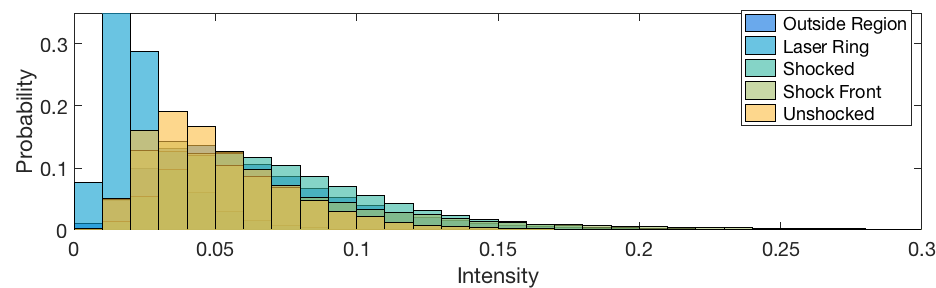}
\caption{(a) An image of a laser-induced, cylindrically converging shock wave propagating in water. 
(b) Training data selected from each class and overlaid on the original image. (c)~Histograms of training data demonstrating overlap between classes throughout image.}
\label{fig:shock}
\end{center}
\end{figure}

When shock waves travel through a material, the material is irreversibly changed. Temporal quantitative analysis of wave dynamics requires all measurements, (in this case, images) to be collected from a single shock experiment. As a result, the signal-to-noise ratio is limited and the images contain high noise content, especially for high spatiotemporal resolution \citep{Kohse}. 
When strong shock waves travel through a material, they can separate into multi-wave structures, beginning with a one-dimensional elastic wave where the material is reversibly compressed and followed by a series of plastic waves corresponding to the irreversible changes that occur upon fast dynamic compression. In an image, the multi-wave structure can appear as a relatively uniform material with bright, narrow attributes that separate the regions. In some cases, there are changes in pixel intensity or in texture between the different regions of the shock, but in most cases these classes are difficult to separate with current boundary detection techniques because the various physics classes (e.g. shock wave) have very little to distinguish them from the other classes (e.g. are overlapping in feature space). 

Figure \ref{fig:shock}(a) is an image of a cylindrically converging shock wave traveling through a thin layer of water that is between thick glass substrates \citep{Pezeril}. The shock wave is generated from the interaction of a 200 micrometer diameter laser ring with an absorber to produce a shock wave that travels within the sample plane, perpendicular to the incident laser. A 180 femtosecond duration pulse from the same laser collects a shadowgraph image in transmission, which gives an image that is a spatial map of the second derivative of the density for the material \citep{Settles}. In this experiment, six images were taken in a single experiment at 5 nanosecond intervals in order to visualize the convergence and subsequent divergence of the shock, which depicts the two-dimensional physics of the complex system. To obtain quantifiable data from these images, the shock must be precisely and accurately located with clear understanding of any error from that measurement. Identifying the location of the shock wave throughout a series of images will enable the researcher to compute quantities such as velocity.

Figure \ref{fig:shock}(b) displays the training data selected for the shock wave image, with the classes beginning at the center of the image and moving out: unshocked water, shock front, shocked water, laser ring, outside region. $85\%$ of the image is represented in the training data, and Figure \ref{fig:shock}(c) displays a histogram of the global training data, of which all classes have significant overlap. Traditional methods fail to provide adequate identification of boundaries on this image, but the underlying cause of this failure is exactly the motivation for why a spatial segmentation technique that adapts to local intensity variations is appropriate for this image.

\begin{figure}[h]
\begin{center}
(a)\includegraphics[width=0.65\textwidth]{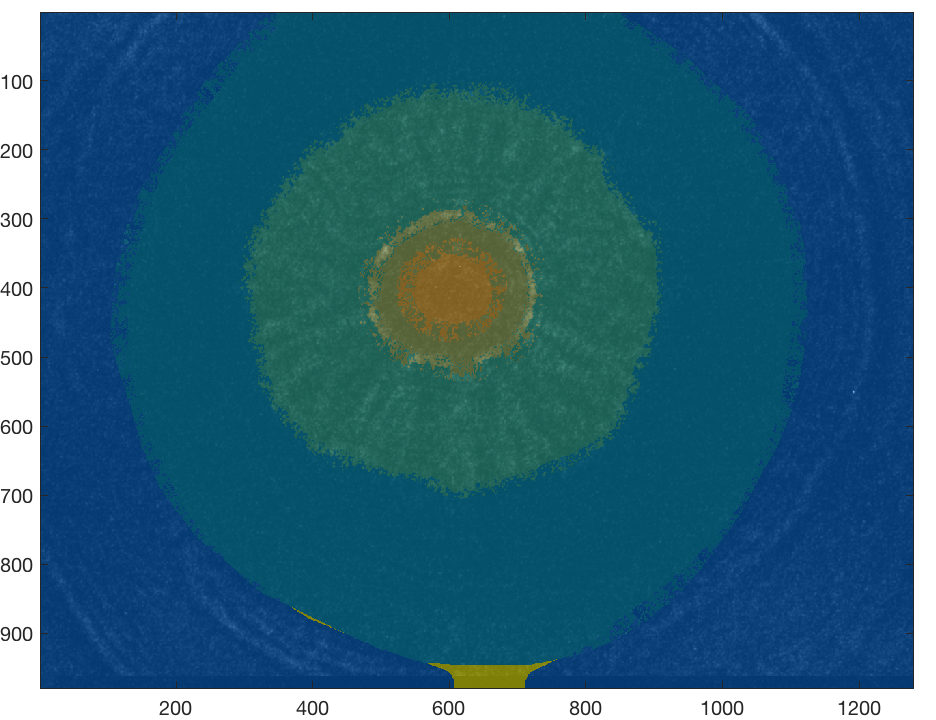} \\
(b)\includegraphics[width=0.45\textwidth]{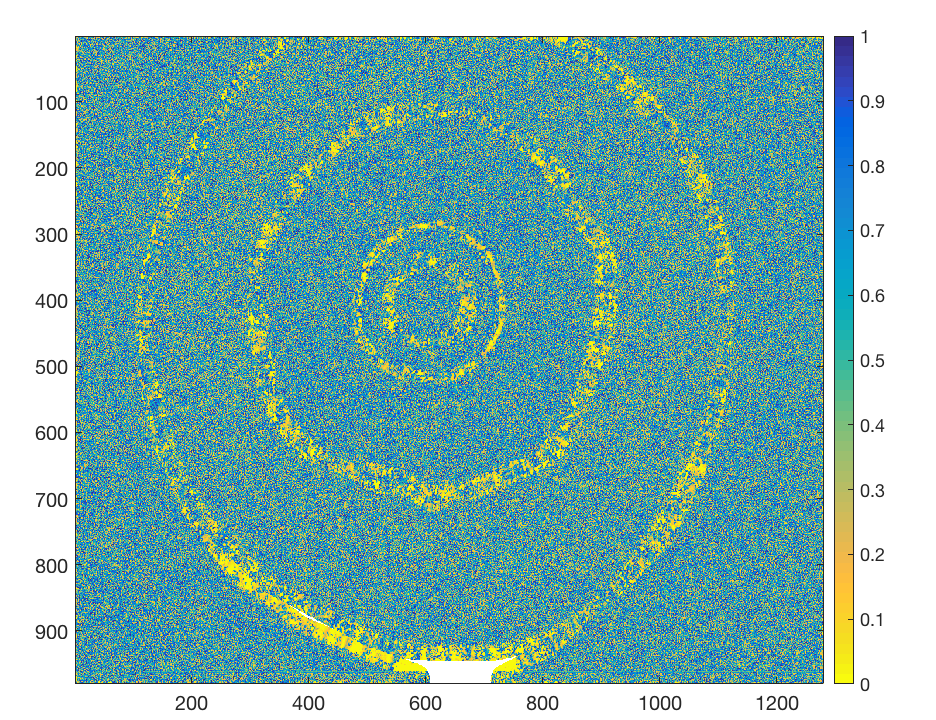}
(c)\includegraphics[width=0.45\textwidth]{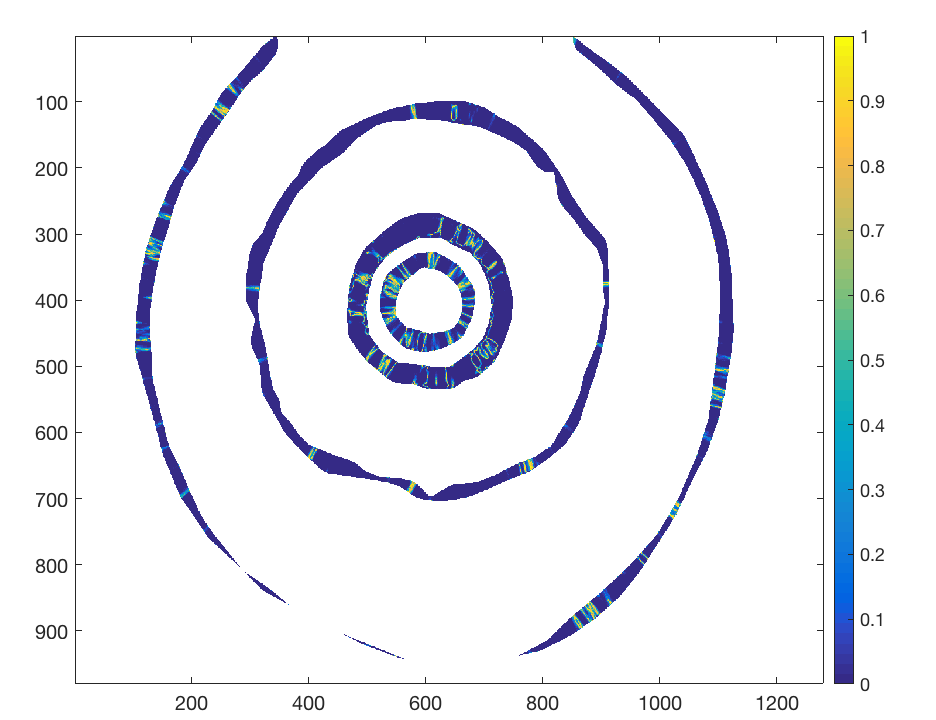}
\caption{(a) The LADA segmentation, with a few pixels identified as bonus class because no local training data were available. (b) The MLE p-value map indicates the regions about which the segmentation has higher uncertainty (low p-value). (c) The ANOVA p-value map indicates some regions in which it was difficult to discriminate between local classes (large p-value).}
\label{fig:shock_seg}
\end{center}
\end{figure}

The LADA segmentation is given in Figure \ref{fig:shock_seg}(a) for $d=25$ and $n=20$, with a smaller $n$ than in the previous example to provide a more narrowed local view of the data. With large gaps in the training data and a small value of $d$, some pixels lack \textit{local} training data and are segmented into the \textit{bonus class}, located near the outer boundary of the laser ring at the bottom of the image. A classical method such as QDA, for comparison, produces a nonsensical segmentation, based on the physics, and is not displayed here. MLE p-values are given in Figure \ref{fig:shock_seg}(b) and indicate regions of uncertainty in the segmentation (bright yellow regions). Note that p-values are not computed for pixels placed in the bonus class. The ANOVA p-values are given in Figure \ref{fig:shock_seg}(c) and indicate regions in which the local classes were difficult to discriminate between (large p-value). Notice the less disparate regions in the ANOVA p-values are radial shapes, not unlike the texture in the image.

\begin{figure}[h]
\begin{center}
\includegraphics[width = .9\textwidth]{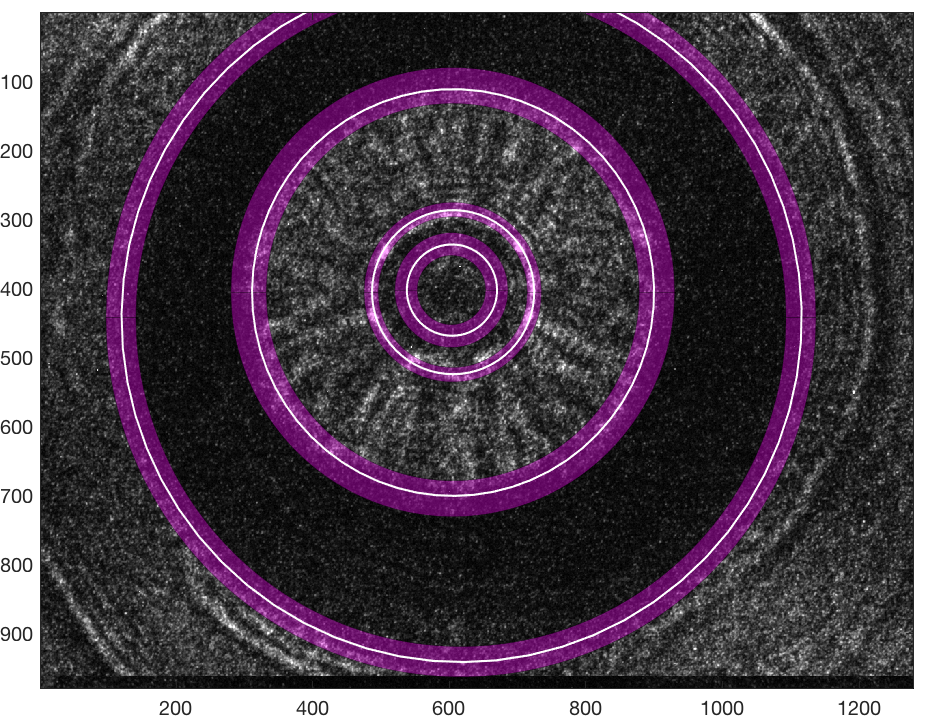}
\caption{The final boundaries (white circle) with uncertainty bounds (purple shade) for both the inner and outer laser ring and the inner and outer shock wave.}
\label{fig:waterUncert}
\end{center}
\end{figure}

With the segmentation in Figure \ref{fig:shock_seg}(a), the boundaries between classes are obtained by taking the gradient. Given circular assumptions on the laser ring and shock fronts, we fit circles to each boundary. To compute bands of uncertainty on the class boundaries, we take a Neyman-Pearson approach and choose a significance level of $\alpha = 0.05$ on the p-value map in Figure \ref{fig:shock_seg}(c). From the p-values $< \alpha$, we determine the smallest and largest radius of uncertainty about each boundary and use that as a uniform uncertainty band about each boundary. Both the fitted circles and the corresponding uncertainty bands are shown in Figure~\ref{fig:waterUncert}, superimposed upon the original image. Note that there are no requirements that the uncertainty band be symmetric about its fitted circular boundary.

\section{Conclusions}
\label{sec:conclusions}
This work presents a locally-adaptive segmentation method for boundary identification, designed for images that contain spatial trends in intensity, low contrast, or heteroskedasticity. Building on the classical image segmentation technique of discriminant analysis, one of our novel contributions is an algorithm that determines the class for each pixel by comparing probability distributions constructed from \textit{local} training data. In recognizing spatial dependence of pixels, we shift from the discriminant analysis paradigm of viewing training data globally and focus on local training data that are restricted by two user-specified parameters. We assume the local training data are Gaussian, which we demonstrate is more reasonable than a global Gaussian assumption on a radiograph produced at the Nevada National Security Site. 

Understanding uncertainty in the segmentation and region boundaries is provided through two statistical tests, the application of which is our second novel contribution. Given a pixel's segmentation and the statistics estimated from local training data, we use maximum likelihood estimation (MLE) to compute the probability of observing the pixel's intensity or something more extreme. For small p-values, this suggests that either the segmentation may not be the true class, even if it was most probable of the locally available classes, or that the local training data are all too far from the pixel to be truly representative. We apply MLE, paired with physics geometric assumptions, to create uncertainty bands for the segmentation boundaries in the radiograph and in an image of a converging shock wave. Finally, an analysis of variance (ANOVA) describes the ability to discriminate between local classes based on the training data, with small p-values indicating well-separable classes. The ANOVA p-value map ultimately describes class contrasts and associated variances and may be used to guide adjusting the local parameters or training data to provide better separability between classes.


\acks{The authors are grateful to Dongsheng Wu at the University of Alabama in Huntsville and Aaron Luttman at the Nevada National Security Site for their advice and support. The authors also acknowledge Keith A. Nelson at MIT for his support in the shock work. This manuscript has been authored in part by National Security Technologies, LLC, under Contract No. DE-AC52-06NA25946 with the U.S. Department of Energy, National Nuclear Security Administration, NA-10 Office of Defense Programs, and supported by the Site-Directed Research and Development Program. The United States Government retains and the publisher, by accepting the article for publication, acknowledges that the United States Government retains a non-exclusive, paid-up, irrevocable, world-wide license to publish or reproduce the published content of this manuscript, or allow others to do so, for United States Government purposes. The U.S. Department of Energy will provide public access to these results of federally sponsored research in accordance with the DOE Public Access Plan (http://energy.gov/downloads/doe-public-access-plan). DOE/NV/25946-{}-3282.} The authors would like to acknowledge the Office of Naval Research for funding that supported one of the authors from this work on grant numbers N00014-16-1-2090 and N00014-15-1-2694.

\vskip 0.2in
\bibliography{LADA_AlgPaper}

\begin{thebibliography}{27}
\providecommand{\natexlab}[1]{#1}
\providecommand{\url}[1]{\texttt{#1}}
\expandafter\ifx\csname urlstyle\endcsname\relax
  \providecommand{\doi}[1]{doi: #1}\else
  \providecommand{\doi}{doi: \begingroup \urlstyle{rm}\Url}\fi

\bibitem[Bazille et~al.(1994)Bazille, Guttman, Mc{V}eigh, and
  Zerhouni]{Bazille}
Anne Bazille, Michael~A. Guttman, Elliot~R. Mc{V}eigh, and Elias~A. Zerhouni.
\newblock Impact of semiautomated versus manual image segmentation errors on
  myocardial strain calculation by magnetic resonance tagging.
\newblock \emph{Invest. Radiol.}, 29:\penalty0 427--433, 1994.

\bibitem[Bishop(2006)]{Bishop}
Christopher~M. Bishop.
\newblock \emph{Pattern Recognition and Machine Learning}.
\newblock Springer, Singapore, 2006.

\bibitem[Bogaert et~al.(2016)Bogaert, Waldner, and Defourny]{Bogaert}
Patrick Bogaert, Fran{\c{c}}ois Waldner, and Pierre Defourny.
\newblock An information-based criterion to measure pixel-level thematic
  uncertainty in land cover classifications.
\newblock \emph{Stoch. Environ. Res. Risk Assess.}, pages 1--16, 2016.

\bibitem[Byng et~al.(1996)Byng, Boyd, Little, Lockwood, Fishell, Jong, and
  Yaffe]{Byng}
J.~W. Byng, N.~F. Boyd, L.~Little, G.~Lockwood, E.~Fishell, R.~A. Jong, and
  M.~J. Yaffe.
\newblock Symmetry of projection in the quantitative analysis of mammographic
  images.
\newblock \emph{Eur. J. Cancer Prev.}, 5:\penalty0 319--327, 1996.

\bibitem[Carpenter et~al.(1999)Carpenter, Gopal, Macomber, Martens, Woodcock,
  and Franklin]{Carpenter}
Gail~A. Carpenter, Sucharita Gopal, Scott Macomber, Siegfried Martens,
  Curtis~E. Woodcock, and Janet Franklin.
\newblock A neural network method for efficient vegetation mapping.
\newblock \emph{Remote Sens. Environ.}, 70:\penalty0 326--338, 1999.

\bibitem[Casella and Berger(2002)]{Casella}
George Casella and Roger~L. Berger.
\newblock \emph{Statistical Inference}.
\newblock Duxbury, Pacific Grove, California, 2nd edition, 2002.

\bibitem[DeGroot and Schervish(2002)]{DeGroot}
Morris~H. DeGroot and Mark~J. Schervish.
\newblock \emph{Probability and Statistics}.
\newblock Addison-Wesley, Boston, 3rd edition, 2002.

\bibitem[Hand(2006)]{Hand}
David~J. Hand.
\newblock Classifier technology and the illusion of progress.
\newblock \emph{Statistical Science}, 21:\penalty0 1--14, 2006.

\bibitem[Haralick(1984)]{Haralick}
Robert~M. Haralick.
\newblock Digital step edges from zero crossing of second directional
  derivatives.
\newblock \emph{IEEE Transactions on Pattern Analysis and Machine
  Intelligence}, PAMI-6\penalty0 (1):\penalty0 58--68, 1984.

\bibitem[Hastie et~al.(2001)Hastie, Tibshirani, and Friedman]{Hastie}
Trevor Hastie, Robert Tibshirani, and Jerome Friedman.
\newblock \emph{{The Elements of Statistical Learning: Data Mining, Inference,
  and Prediction}}.
\newblock Springer-Verlag, New York, 2nd edition, 2001.

\bibitem[Hell et~al.(2015)Hell, Kassubeck, Bauszat, Eisermann, and
  Magnor]{Hell}
Benjamin Hell, Marc Kassubeck, Pablo Bauszat, Martin Eisermann, and Marcus
  Magnor.
\newblock An approach toward fast gradient-based image segmentation.
\newblock \emph{IEEE Trans. Image Process.}, 24:\penalty0 2633--2645, 2015.

\bibitem[Howard et~al.(2016)Howard, Fowler, Luttman, Mitchell, and
  Hock]{Howard16}
Marylesa Howard, Michael Fowler, Aaron Luttman, Stephen~E. Mitchell, and
  Margaret~C. Hock.
\newblock Bayesian abel inversion in quantitative x-ray radiography.
\newblock \emph{J. Comput. Appl. Math.}, 38:\penalty0 B396--B413, 2016.

\bibitem[Khatami et~al.(2017)Khatami, Mountrakis, and Stehman]{Khatami}
Reza Khatami, Giorgos Mountrakis, and Stephen~V. Stehman.
\newblock Mapping per-pixel predicted accuracy of classified remote sensing
  images.
\newblock \emph{Remote Sens. Environ.}, 191:\penalty0 156--167, 2017.

\bibitem[Kittler(1983)]{Kittler}
J.~Kittler.
\newblock On the accuracy of the {S}obel edge detector.
\newblock \emph{Image and Vision Comput.}, 1:\penalty0 37--42, 1983.

\bibitem[Kohse-H{\"o}inghaus and Jeffries(2002)]{Kohse}
Katharina Kohse-H{\"o}inghaus and Jay~B. Jeffries.
\newblock \emph{Applied Combustion Diagnostics}.
\newblock CRC Press, New York, 2002.

\bibitem[Mansoor et~al.(2015)Mansoor, Bagci, Foster, Zu, Papadakis, Folio,
  Udupa, and Mollura]{Mansoor}
Awais Mansoor, Ulas Bagci, Brent Foster, Ziyue Zu, Georgios~Z. Papadakis,
  Les~R. Folio, Fayaram~K. Udupa, and Daniel~F. Mollura.
\newblock Segmentation and image analysis of abnormal lungs at {CT}: Current
  approaches, challenges, and future trends.
\newblock \emph{Radiographics}, 35:\penalty0 1056--1076, 2015.

\bibitem[McIver and Friedl(2001)]{McIver}
Douglas~K. McIver and Mark~A. Friedl.
\newblock Estimating pixel-scale land cover classification confidence using
  nonparametric machine learning methods.
\newblock \emph{IEEE Trans. Geosci. Remote Sens.}, 39:\penalty0 1959--1968,
  2001.

\bibitem[Melin et~al.(2014)Melin, Gonzalez, Castro, Mendoza, and
  Castillo]{Melin}
Patricia Melin, Claudia~I. Gonzalez, Juan~R. Castro, Olivia Mendoza, and Oscar
  Castillo.
\newblock Edge detection method for image processing based on generalized
  type-2 fuzzy logic.
\newblock \emph{IEEE Trans. Fuzzy Systems}, 22:\penalty0 1515--1525, 2014.

\bibitem[Moriwaki et~al.(2014)Moriwaki, Terada, Kose, Haishi, and
  Sekozawa]{Moriwaki}
Toshi Moriwaki, Yasuhiko Terada, Katsumi Kose, Tomoyuki Haishi, and Yoshihiko
  Sekozawa.
\newblock Visualization and quantification of vascular structure of fruit using
  magnetic resonance microimaging.
\newblock \emph{Appl. Magn. Reson.}, 45:\penalty0 517--525, 2014.

\bibitem[Pal and Pal(1993)]{Pal}
Nikhil~R. Pal and Sankar~K. Pal.
\newblock A review on image segmentation techniques.
\newblock \emph{Pattern Recognit.}, 26:\penalty0 1277--1294, 1993.

\bibitem[Pezeril et~al.(2011)Pezeril, Saini, Veysset, Kooi, Fidkowski,
  Radovitzky, and Nelson]{Pezeril}
Thomas Pezeril, Gagan Saini, David Veysset, Steve Kooi, Piotr Fidkowski, Raul
  Radovitzky, and Keith~A. Nelson.
\newblock Direct visualization of laser-driven focusing shock waves.
\newblock \emph{Phys. Rev. Lett.}, 106\penalty0 (21):\penalty0 214--503, 2011.

\bibitem[Plaza et~al.(2012)Plaza, Scheffer, and Saunders]{Plaza}
Stephen~M. Plaza, Louis~K. Scheffer, and Mathew Saunders.
\newblock Minimizing manual image segmentation turn-around time for neuronal
  reconstruction by embracing uncertainty.
\newblock \emph{PLoS One}, 7\penalty0 (9):\penalty0 1--14, 09 2012.

\bibitem[Settles(2001)]{Settles}
Gary~S. Settles.
\newblock \emph{Schlieren and Shadowgraph Techniques: Visualizing Phenomena in
  Transparent Media}.
\newblock Springer, Berlin, 2001.

\bibitem[Woodcock(2002)]{Woodcock}
Curtis~E. Woodcock.
\newblock \emph{Uncertainty in Remote Sensing}, chapter~2, pages 19--24.
\newblock Wiley, West Sussex, 2002.

\bibitem[Xie and Tu(2015)]{Xie}
Saining Xie and Zhuowen Tu.
\newblock Holistically-nested edge detection.
\newblock In \emph{Proc. IEEE Int. Conf. on Comput. Vision}, pages 1395--1403,
  2015.

\bibitem[Zha et~al.(2003)Zha, Gao, and Ni]{Zha}
Yong Zha, Jay Gao, and Shaoxiang Ni.
\newblock Use of normalized difference built-up index in automatically mapping
  urban areas from {TM} imagery.
\newblock \emph{Int. J. Remote Sens.}, 24:\penalty0 583--594, 2003.

\bibitem[Zhao et~al.(2011)Zhao, Stein, Chen, and Zhang]{Zhao}
Xi~Zhao, Alfred Stein, Xiaoling Chen, and Xiang Zhang.
\newblock Quantification of extensional uncertainty of segmented image objects
  by random sets.
\newblock \emph{IEEE Trans. Geosci. Remote Sens.}, 49:\penalty0 2548--2557,
  2011.

\end{thebibliography}

\end{document}